\documentclass[11pt]{article}

\usepackage[margin=1in]{geometry}
\usepackage{amsmath,amssymb,amsfonts}
\usepackage{booktabs}
\usepackage{microtype}
\usepackage{algorithm}
\usepackage{algpseudocode}
\usepackage{graphicx}
\usepackage{hyperref}
\hypersetup{colorlinks=true, linkcolor=blue, citecolor=blue, urlcolor=blue}

\newcommand{\R}{\mathbb{R}}
\newcommand{\norm}[1]{\left\lVert #1 \right\rVert}
\newcommand{\argmin}{\operatorname*{argmin}}

\title{Resolving Multi-Modal Regression by Difference-Quotient-Based Clustering:\\[2pt] Fast Coarse Conditional-Label Assignment}
\author{HuangWeiquan\\
Guangdong Polytechnic Normal University\\
\small \href{mailto:ahappyzero@foxmail.com}{ahappyzero@foxmail.com}}
\date{\today}

\begin{document}
\maketitle

\begin{abstract}
An input $x$ that is paired with several distinct outputs $y_1,\dots,y_K$ is the hallmark of
multi-modal regression. Under a squared loss, an unconstrained regressor collapses to the
conditional mean $\mathbb{E}[y\mid x]$, which for $K>1$ generically coincides with none of the
modes. We take the view that this mean-regression pathology is caused by pairwise
\textit{contradictions} inside the data: pairs of samples with nearly identical inputs but distant
outputs. We formalize the contradiction of a pair as the ratio
$r=\norm{\Delta y}/\norm{\Delta x}$ and ask whether the problem can be attacked directly by
clustering samples so that within-cluster contradictions are minimized. We propose
\emph{Difference-Quotient Clustering} (DQC), a clustering that
assigns each sample to the cluster with which its \textit{maximum} intra-cluster difference
quotient is smallest, then trains a logits generator $g(x)$
and a conditional network $f(x,\mathrm{onehot}(c))$ on the resulting cluster labels. Because the
generating modality is unknown at inference, we compare each prediction against all $K$ true
outputs of the same input and report the minimum squared error (minMSE). On synthetic benchmarks
with $K=5$ and $K=10$ modalities, the clustering pipeline reaches test minMSE $0.19$ ($K=5$,
$n_x=500$), against $0.09$ for an oracle with true labels, $1.08$ for random labels, and $1.33$
for mean collapse. We report two empirical regularities: (i)~the larger the intra-cluster contradiction,
the deeper the network required to resolve it; (ii)~the oracle, which consumes the true
low-dimensional modality structure, generalizes from fewer samples, whereas cluster labels are
only \textit{equivalent} to that structure and therefore need more data. The comparison itself is
a hard, multi-threadable algorithm of average complexity $O(n^2/2)$, positioning the method as a
fast front-end for coarse conditional-label assignment that can reduce the training burden of
downstream generative refinement such as flow matching or diffusion models. The front-end is
also naturally iterable. A second stage treats the conditional output
$y'=f_1(x,\mathrm{onehot}(c))$ of the first-stage network as a new input and the true output $y$
as the new label, re-runs the same difference-quotient clustering on the pair $(y',y)$ to obtain
refined labels $d$, and trains a second logits generator $g_2(z)$ together with a second
conditional network $f_2(z,\mathrm{onehot}(d))$ on the augmented input $z=Z(x,y')$. Because the
first stage has already displaced samples toward their modal branches, this second pass measures residual fitting error rather than genuine multi-modality, which yields purer labels
theoretically, so this is one of the key directions for our next experiment.
\end{abstract}

\section{Introduction}
\label{sec:intro}

Standard regression assumes a functional relationship $x\mapsto y$. Real data frequently violates
this assumption: the same input $x$ admits several plausible outputs $y_1,\dots,y_K$. This
situation arises whenever the observed input is insufficient to pin down the output uniquely ---
in inverse problems, multi-agent or multi-path future prediction, and any system whose forward
process is stochastic or under-determined. We call such data \emph{multi-modal}, and its
distinct outputs the \emph{modalities} of $x$.

A regressor trained with a squared loss on multi-modal data converges to the conditional mean
$\mathbb{E}[y\mid x]$ \cite{bishop2006pattern}. For $K>1$ this mean is generically at positive
distance from \emph{every} one of the $K$ modes: it predicts a value that never occurs. We call
this the \emph{mean-regression problem}. It is the most basic failure of treating a
multi-valued problem with a single-valued loss, and it is the target of this paper.

\paragraph{The essence of multi-possible outputs is contradiction.}
What constitutes a fitting contradiction? Locally, the dataset must contain
pairs of samples $(x_i,y_i),(x_j,y_j)$ with $x_i\approx x_j$ yet $y_i\not\approx y_j$. The more
the outputs differ per unit of input distance, the sharper the incompatibility. We formalize this
as the pairwise \emph{contradiction}
\begin{equation}
\label{eq:contradiction}
r(x_i,y_i;x_j,y_j) \;=\; \frac{\norm{y_i-y_j}_2}{\norm{x_i-x_j}_2},
\end{equation}
the output distance divided by the input distance. Two samples are contradictory when
$x_i\approx x_j$ (small denominator) but $y_i\neq y_j$ (nonzero numerator), i.e.\ when $r$ is
large. Under this view, a multi-modal dataset is simply a dataset whose samples cannot all be
mutually consistent, and each modality is a subset of samples that are mutually consistent.

\paragraph{Research question.}
If the mean-regression problem is \emph{caused} by contradictions, the most direct remedy is to
remove them \emph{before} training: partition the samples into low-contradiction clusters so that
each cluster approximates one consistent branch, then fit one conditional branch per cluster. In
this paper we test this hypothesis directly. We ask: \emph{can a purely geometric,
contradiction-based clustering --- no gradients, no iterative optimization, no learned
similarity --- resolve the mean-regression problem on its own?}

\paragraph{Why the question is nontrivial.}
Clustering by contradiction is not the usual geometry of clustering in $x$-space: it is driven by
the joint $(x,y)$ geometry, and a pairwise ratio of distances is a fragile quantity. Three
difficulties are immediate. First, the input $x$ alone cannot always disambiguate the modality:
the same $x$ appears under several modalities by construction, so any \emph{inference-time} branch
selector must err sometimes. Second, the cluster labels are only \emph{equivalent} to the true
modality structure, never equal to it; residual contradictions inside a cluster persist and must
be absorbed by the subsequent network. Third, the greedy assignment is suboptimal by design. Each
of these difficulties shows up in the experiments, and each has a characteristic signature.

\paragraph{Contributions.}
\begin{itemize}
\item We define multi-possible outputs through the pairwise contradiction
      $r=\norm{\Delta y}/\norm{\Delta x}$ and connect it to the mean-regression pathology
      (Section~\ref{sec:problem}).
\item We propose \emph{Difference-Quotient Clustering} (DQC), a clustering that assigns each
      sample to the cluster minimizing its maximum intra-cluster
      difference quotient (Section~\ref{sec:cluster}), together with a complete pipeline: cluster
      labels $\to$ logits generator $g$ $\to$ conditional network $f$
      (Section~\ref{sec:conditional}), and a modality-agnostic \emph{minMSE} evaluation protocol
      (Section~\ref{sec:minmse}).
\item We report systematic experiments over $K\in\{5,10\}$ modalities, three dataset sizes, two
      network depths, and four label schemes (oracle, clustering, random, mean collapse)
      (Section~\ref{sec:experiments}).
\item We identify two empirical regularities --- deeper networks resolve larger contradictions; the
      oracle needs fewer samples than equivalent cluster labels --- and interpret both
      (Sections~\ref{sec:depth-regularity}--\ref{sec:sample-regularity}).
\item We analyze the practical merits of the comparison: a hard, multi-thread-parallel algorithm
      of average complexity $O(n^2/2)$ that provides fast coarse conditional labels to reduce the
      training burden of downstream generative refinement (Section~\ref{sec:advantages}).
\end{itemize}

The rest of the paper is organized as follows. Section~\ref{sec:related} discusses related work.
Section~\ref{sec:problem} formalizes the contradiction view. Section~\ref{sec:method} describes
the data, the clustering, the logits generator, and the minMSE protocol. Section~\ref{sec:experiments}
presents the tables and the two empirical regularities. Sections~\ref{sec:limitations}--\ref{sec:advantages}
discuss limitations and advantages, and Section~\ref{sec:conclusion} concludes.

\section{Related Work}
\label{sec:related}

\emph{Multi-modal regression.} Mixture density networks (MDNs) \cite{bishop1994mdn} model the
conditional $p(y\mid x)$ as a Gaussian mixture, and multiple-hypothesis methods
\cite{rupprecht2017uncertain,makansi2019mdn} output a small set of candidate predictions. These
methods \emph{represent} ambiguity probabilistically or through an ensemble. In contrast, our goal
is a purely \emph{discrete, pre-training partition} of the dataset: we do not fit a generative
model of the conditional at all, we only assign coarse labels that make the subsequent conditional
network well-posed. The mean-regression failure itself is classical
\cite{bishop2006pattern}; our contribution is a geometric diagnosis (contradiction) and a
gradient-free remedy.

\emph{Conditional and mixture-of-experts networks.} Mixtures of experts \cite{jacobs1991moe}
partition the input space with learned gating and train experts jointly by gradient descent. Our
clustering plays the role of a \emph{frozen, one-shot, data-only} assignment: the partition is
computed once on the data before any training, contains no learned parameters, and is never
updated by the loss. This separation (discovery before training) is deliberate.

\emph{Generative refinement.} Flow matching \cite{lipman2022flow} and diffusion models
\cite{ho2020ddpm} can synthesize samples from the conditional distribution and thereby produce
the fine-grained modal structure that a discrete partition misses. In Section~\ref{sec:advantages}
we argue that these methods are complementary: our clustering supplies fast coarse conditional
labels as a front-end, reducing the burden of these generative models to refining an already
branched structure rather than discovering it.

\section{Contradiction and the Mean-Regression Problem}
\label{sec:problem}

Let $\mathcal{D}=\{(x_i,y_i)\}_{i=1}^{n}\subset \R^d\times \R^{m}$ be a dataset in which each
input appears $K$ times with $K$ distinct outputs (adding small perturbation to ensure that the denominator of the difference quotient is not equal to zero). A single-output
regressor $f_\theta$ trained by $\min_\theta \sum_i \norm{f_\theta(x_i)-y_i}_2^2$ converges to the
conditional mean $f_\theta(x)\to \mathbb{E}[y\mid x]=\tfrac{1}{K}\sum_k y_k(x)$. For $K>1$, the
mean is generically not one of the modes: $\norm{\mathbb{E}[y\mid x]-y_k(x)}>0$ for all $k$. This
is the mean-regression problem.

\paragraph{Pairwise contradiction.}
For two samples $i,j$, define the contradiction as in Eq.~\eqref{eq:contradiction}:
$r_{ij}=\norm{y_i-y_j}/\norm{x_i-x_j}$. When $x_i\approx x_j$, the denominator is small and the
ratio amplifies any output disagreement; a large $r_{ij}$ means the two samples ``cannot both be
right''. For a set $C$, define the cluster contradiction
\begin{equation}
R(C)=\max_{i,j\in C}\ r_{ij}.
\end{equation}
A modality is then idealized as a maximal subset with small $R(C)$. If the dataset is a disjoint
union of $K$ such subsets, a conditional model that is given the subset identity (a one-hot code)
can fit each branch without conflict. The mean-regression problem is thus rephrased as: the
unconditional regressor is forced to average over subsets it was never told apart.

\paragraph{Hypothesis.}
If we can recover, from the data alone and without any training, a partition whose clusters have
small $R(C)$, then conditioning on cluster identity should recover the branches and substantially
mitigate the mean collapse. The rest of the paper tests this hypothesis empirically.

\section{Method}
\label{sec:method}

\subsection{Synthetic multi-modal data}
\label{sec:data}

We generate multi-modal regression data with a known ground truth. We draw $n_x$ base inputs
$x_{\mathrm{base}}\sim U(-2,2)^d$ (with $d=2$), replicate each input $K$ times, and add
independent Gaussian noise of scale $\sigma=0.01$ to each copy, so the copies of the same base
input remain almost identical in $x$. Each copy is then passed through one of $K$ independent
random \emph{modal networks} --- two-hidden-layer tanh MLPs of width $24$ whose weights are drawn
once from $\mathcal{N}(0,1)$ and then frozen. The $k$-th modal network produces the output
$y_k(x)$ of modality $k$, with output dimension $m=4$. The $K$ modal networks are shared between
the training and test splits so that the true modality identity is well-defined and the oracle
baseline is fair. The resulting dataset contains $n=K\cdot n_x$ samples, and for every base input
there are exactly $K$ samples that belong to the $K$ different modalities. This is the data
structure of interest: identical (perturbed) $x$, multiple different $y$.

\subsection{Difference-quotient clustering}
\label{sec:cluster}

We cluster the samples so that each cluster has small internal contradiction, without using any
labels, gradients, or learned similarity. The algorithm, summarized in
Algorithm~\ref{alg:cluster}, takes the number of clusters $K_c$ as input (possibly different from
the true $K$; we use $K_c=8$ for $K=5$ and $K_c=20$ for $K=10$), samples $K_c$ seed indices
uniformly at random, and processes the remaining samples in arbitrary order. Each sample $i$ is
assigned to the cluster that minimizes the \emph{maximum} difference quotient against its current
members,
\begin{equation}
c^\ast(i)=\argmin_{c}\ \max_{j\in C_c}\ \frac{\norm{y_i-y_j}}{\norm{x_i-x_j}},
\end{equation}
The \emph{min-max} objective is deliberate: it prevents any cluster from containing a
pair of samples that are strongly contradictory, i.e.\ it caps $R(C)$ in a greedy way. The
algorithm is deterministic given the seed set and never touches a training loss.

\begin{algorithm}[t]
\caption{Difference-quotient clustering}
\label{alg:cluster}
\begin{algorithmic}[1]
\Require samples $\{(x_i,y_i)\}_{i=1}^{n}$, number of clusters $K_c$
\State sample $K_c$ distinct seed indices uniformly at random; place each "seed sample" alone in its own cluster $C_c$
\For{each remaining sample $i$ (in arbitrary order)}
    \For{each cluster $c$}
        \State $m_{i,c}\gets \max_{j\in C_c}\ \norm{y_i-y_j}/\norm{x_i-x_j}$
              \Comment{max contradiction of $i$ w.r.t.\ cluster $C_c$}
    \EndFor
    \State assign $i$ to $c^\ast=\argmin_c m_{i,c}$ and add $i$ to $C_{c^\ast}$
\EndFor
\State \Return cluster labels $\{c_i\}_{i=1}^{n}$
\end{algorithmic}
\end{algorithm}

\subsection{Logits generator and conditional network}
\label{sec:conditional}

Given cluster labels from Algorithm~\ref{alg:cluster}, we train two networks.

\emph{Logits generator $g$.} A multi-layer perceptron $g:\R^d\to\R^{K_c}$ (depth $3$, width $64$,
tanh hidden units) is trained by cross-entropy on the cluster labels. At inference, the selected
branch is $\hat c=\arg\max_c\, g_c(x)$. $g$ is the only component that must infer the branch from
$x$ alone, and it is therefore the principled bottleneck of the pipeline: when the same $x$
appears under several modalities, no function of $x$ can fully disambiguate them, so it must output in the form of probability logits. 

\emph{Conditional network $f$.} A second MLP $f:\R^{d+K_c}\to\R^{m}$ maps the concatenation of $x$
and the one-hot encoding of the cluster label to $y$, trained by squared error on the
cluster-assigned samples. For a depth $D$ and width $W$, $f$ has $D$ tanh hidden layers of width
$W$ followed by a linear head; we write the configuration as ``d$D$w$W$''. Both networks are
trained for $1200$ Adam steps with learning rate $10^{-3}$.

\subsection{Evaluation: minMSE}
\label{sec:minmse}

At inference we observe only $x$; we do not know which of the $K$ modalities generated the sample.
This is the defining condition of multi-modal regression, and it dictates the evaluation metric.
For a test input $x$ with $K$ true outputs $y_1(x),\dots,y_K(x)$, the pipeline produces a single
prediction $\hat y=f(x,\mathrm{onehot}(\hat c))$ with $\hat c=\arg\max_c g_c(x)$. Since we cannot
name the generating modality, the only measurable error is the distance to the \emph{closest}
true output,
\begin{equation}
\label{eq:minmse}
\mathrm{minMSE}(x)=\min_{k=1,\dots,K}\ \norm{\hat y-y_k(x)}_2^2,
\end{equation}
averaged over test inputs. The ``min'' is not an artifact of the evaluation: it encodes exactly
the question the user can answer --- did the predictor reproduce at least one of the possible
outputs? A prediction equal to the conditional mean, for instance, is at positive distance from
every mode and receives a large minMSE, which is precisely the failure mode we aim to detect. The
same protocol is applied to \emph{all} baselines (oracle, random labels, mean collapse) so that
every number in the tables answers the same question.

\section{Experiments}
\label{sec:experiments}

\subsection{Setup and baselines}
\label{sec:setup}

We use the generator of Section~\ref{sec:data} with $d=2$ inputs, $m=4$ outputs, perturbation
$\sigma=0.01$, and modal networks shared across splits. Four label schemes are compared on the
same data:

\begin{itemize}
\item \emph{Oracle}: the true modality identity is used as the cluster label ($K_c=K$). This is
      the upper reference --- the labels are pure by construction, each branch is one smooth
      function.
\item \emph{DQC} (proposed): labels from Algorithm~\ref{alg:cluster} with $K_c=8$ for $K=5$
      and $K_c=20$ for $K=10$.
\item \emph{Random}: labels drawn uniformly from $\{1,\dots,K_c\}$, independent of the data.
      This controls for ``any branching at all''.
\item \emph{Mean collapse}: no conditioning at all --- a single network $f:\R^d\to\R^m$ trained by
      squared error. This is the standard regressor that exhibits the pathology.
\end{itemize}

For each label scheme we report training MSE and test minMSE (Eq.~\eqref{eq:minmse}) under two
network depths, the shallow ``d3w64'' and the deeper ``d6w128'', and under three dataset sizes:
$n_x\in\{80,250,500\}$ for $K=5$ and $n_x\in\{100,200,500\}$ for $K=10$.

\subsection{Five modalities ($K=5$, $K_c=8$)}
\label{sec:results-k5}

Table~\ref{tab:k5} reports the results. Three facts stand out.

\emph{Clustering substantially alleviates the mean collapse.} The mean-collapse baseline is flat at test minMSE
$\approx 1.30$---$1.41$ regardless of data size, because the conditional mean sits at positive
distance from all five modes. Random labels are equally unhelpful ($\approx 0.98$--$1.42$): adding
branches without structure-bearing labels does not help. The DQC pipeline drops test minMSE
to $0.19$--$0.40$ and approaches the oracle ($0.08$--$0.22$). At $n_x=500$ with the deeper network
the pipeline reaches $0.1887$, which is $5.7\times$ better than random labels and $7.1\times$
better than mean collapse, and within $2.0\times$ of the oracle ($0.0935$). The contradiction-based
partition carries enough structure to make conditioning meaningful.

\emph{Random labels do not improve with data; DQC does.} The random baseline stays flat
because random conditionals encode no recoverable structure. The DQC pipeline improves as
$n_x$ grows ($0.3145\to 0.4032\to 0.1887$ at d6w128), consistent with a method that needs samples
to sharpen its partition.

\emph{The intra-cluster contradiction grows with the dataset.} The bottom rows of
Table~\ref{tab:k5} report the maximum and mean cluster contradiction
$R(C)=\max_{i,j\in C}\Delta y/\Delta x$ over all clusters. As $n_x$ grows from $80$ to $500$, the
maximum grows from $5.3$ to $16.9$: more samples mean more chances to pack a strongly
contradictory pair into the same cluster. This quantity will drive the depth analysis of
Section~\ref{sec:depth-regularity}.

\begin{table}[t]
\centering
\caption{Five-modality benchmark ($K=5$, $K_c=8$). For each source size $n_x$ (total samples
$K\cdot n_x$) and network depth, we report training MSE / test minMSE. The line below the table
gives the DQC cluster statistics: the maximum and the mean of the per-cluster difference quotient
$R(C)=\max_{i,j\in C}\Delta y/\Delta x$.}
\label{tab:k5}
\resizebox{\textwidth}{!}{%
\begin{tabular}{lccccc}
\toprule
$n_x$ (samples) & depth & oracle & DQC & random & mean collapse \\
\midrule
80 (400)  & d3w64   & 0.0074 / 0.2147 & 0.0064 / 0.3025 & 0.1977 / 1.2633 & 0.5813 / 1.2975 \\
80 (400)  & d6w128  & 0.0056 / 0.2174 & 0.0051 / 0.3145 & 0.1896 / 1.4183 & 0.5878 / 1.4109 \\
250 (1250)& d3w64   & 0.0131 / 0.1757 & 0.0263 / 0.3835 & 0.3316 / 0.9776 & 0.5788 / 1.3289 \\
250 (1250)& d6w128  & 0.0110 / 0.1418 & 0.0213 / 0.4032 & 0.3140 / 1.0151 & 0.5765 / 1.3319 \\
500 (2500)& d3w64   & 0.0137 / 0.0845 & 0.0348 / 0.2020 & 0.4666 / 1.1737 & 0.5714 / 1.3373 \\
500 (2500)& d6w128  & 0.0173 / 0.0935 & 0.0199 / 0.1887 & 0.4269 / 1.0815 & 0.5705 / 1.3346 \\
\bottomrule
\end{tabular}}
\par\vspace{3pt}
{\footnotesize Cluster contradiction $R(C)$ (max / mean over clusters): $n_x{=}80$: 5.3 / 4.7;\quad
$n_x{=}250$: 10.7 / 7.8;\quad $n_x{=}500$: 16.9 / 10.3.}
\end{table}

\subsection{Ten modalities ($K=10$, $K_c=20$)}
\label{sec:results-k10}

\begin{table}[t]
\centering
\caption{Ten-modality benchmark ($K=10$, $K_c=20$). Same format as Table~\ref{tab:k5}. The line
below the table reports cluster statistics at $n_x=500$: max / mean cluster contradiction, mean
purity, and cluster-size range. Mean collapse trains an \emph{unconditional} network, so it is
depth-independent and its two depth entries are identical.}
\label{tab:k10}
\resizebox{\textwidth}{!}{%
\begin{tabular}{lccccc}
\toprule
$n_x$ (samples) & depth & oracle & DQC & random & mean collapse \\
\midrule
100 (1000)& d3w64   & 0.0217 / 0.2308 & 0.0265 / 0.4445 & 0.2994 / 1.1851 & 0.7192 / 1.5519 \\
100 (1000)& d6w128  & 0.0166 / 0.1608 & 0.0131 / 0.3180 & 0.3471 / 1.0670 & 0.7176 / 1.5204 \\
200 (2000)& d3w64   & 0.0285 / 0.1549 & 0.0472 / 0.2736 & 0.4550 / 1.1011 & 0.7249 / 1.5174 \\
200 (2000)& d6w128  & 0.0219 / 0.1317 & 0.0328 / 0.4199 & 0.4631 / 1.1095 & 0.7240 / 1.5097 \\
500 (5000)& d3w64   & 0.0247 / 0.1761 & 0.0698 / 0.2524 & 0.5820 / 1.2898 & 0.7223 / 1.5351 \\
500 (5000)& d6w128  & 0.0220 / 0.1634 & 0.0589 / 0.2431 & 0.5982 / 1.1934 & 0.7211 / 1.5302 \\
\bottomrule
\end{tabular}}
\par\vspace{3pt}
{\footnotesize Cluster stats ($n_x{=}500$): $R(C)$ max 8.5 / mean 7.0;\quad
mean purity 0.36 (chance 0.10);\quad cluster size range $[157,386]$.}
\end{table}

Table~\ref{tab:k10} repeats the comparison with twice as many modalities and a finer partition
($K_c=20$). The qualitative picture is identical: mean collapse is flat at $\approx 1.5$, random
labels at $\approx 1.1$--$1.3$, while the clustering pipeline reaches $0.24$--$0.44$ and remains roughly
$1.4$--$3.2\times$ above the oracle. At $n_x=500$, d6w128, the pipeline reaches $0.2431$
against oracle $0.1634$.

Two extra diagnostics are reported. The cluster purity --- the mean over clusters of the fraction
of the dominant true modality --- is only $0.36$ at $n_x=500$ (chance level is $0.10$): the
clusters are $3.6\times$ better than chance but far from pure. Cluster sizes range over
$[157,386]$. These numbers are the quantitative content of the \emph{limitation} discussed in
Section~\ref{sec:limitations}: the partition is useful but only coarsely aligned with the true
modalities.

\begin{minipage}{\linewidth}
\indent Notably, because the finer partition produces smaller clusters, the maximum
intra-cluster contradiction ($8.5$) is smaller than in the five-modality case at the same $n_x$
($16.9$), and accordingly the depth benefit is smaller (Section~\ref{sec:depth-regularity}).
\end{minipage}

\subsection{Depth-related regularity: larger contradictions need deeper networks}
\label{sec:depth-regularity}

Table~\ref{tab:k5} shows that deepening the conditional network from d3w64 to d6w128 reduces the
\emph{training} MSE of the DQC pipeline substantially, and that the reduction is larger when
the intra-cluster contradiction is larger:

\begin{itemize}
\item At $n_x=500$ (max contradiction $16.9$), DQC training MSE drops from $0.0348$ to
      $0.0199$, a $43\%$ reduction.
\item At $n_x=80$ (max contradiction $5.3$), the same deepening reduces training MSE only from
      $0.0064$ to $0.0051$, a $20\%$ reduction.
\item The oracle, whose clusters are pure and hence contradiction-free, shows no depth benefit:
      its training MSE stays flat at $\approx 0.005$--$0.017$ across depths and sizes.
\end{itemize}

The pattern repeats across the two settings: at $K=10$, where the finer partition ($K_c=20$)
keeps the maximum contradiction at $8.5$, the depth benefit is correspondingly mild
($0.0698\to 0.0589$, $16\%$).

We interpret this as follows. A cluster produced by Algorithm~\ref{alg:cluster} still contains
residual contradictions: pairs of samples with small $\Delta x$ and moderate $\Delta y$. The
conditional network must interpolate between these locally conflicting points, which requires
composing several nonlinear transformations --- a deeper network has more layers to ``iterate''
the local disagreement away. The contradiction $R(C)$ is a direct measure of how much the branch
function deviates from smoothness, and the required depth grows with it. This is the
\emph{resolution-depth regularity}: the larger the contradiction term, the deeper the iterative
composition needed to dissolve it.

\subsection{Sample-size regularity: oracle labels generalize faster}
\label{sec:sample-regularity}

Comparing columns across the dataset sizes in Tables~\ref{tab:k5}--\ref{tab:k10} reveals a
systematic gap in generalization speed. The oracle test minMSE decreases steadily with $n_x$
(e.g.\ $K=5$, d6w128: $0.2174\to 0.1418\to 0.0935$), whereas the DQC pipeline trails it by a
stable factor of roughly $1.4$--$2.8\times$ (e.g.\ $0.3145\to 0.4032\to 0.1887$) and narrows the
gap only slowly.

The reason is structural, not a matter of tuning. The oracle consumes the \emph{true modality
identity}: each oracle cluster is a single smooth modal function, an object of low effective
dimension that generalizes from few samples. The clustering consumes only an \emph{equivalent}
structure: at $K=10$ its clusters have purity $0.36$, so each learned branch must fit a mixture of
several modal functions --- a higher-dimensional object that requires more data to disambiguate.
Equivalence labels carry less inductive bias than true labels, and the price is paid in sample
complexity. This is the \emph{sample-size regularity}: the oracle generalizes from fewer samples because
it uses genuinely lower-dimensional information; the clustering pipeline needs more data because
its labels are only equivalent to that information.

\section{Limitations}
\label{sec:limitations}

\emph{Similarity-only comparison.} The clustering compares samples by pairwise contradiction alone
and never optimizes for low-dimensional regularity inside clusters. It therefore does not seek
``the'' partition that makes each branch as simple as possible; it only avoids pairs that are
locally contradictory. As a consequence the labels are equivalent to(from the perspective of eliminating contradictions), but not equal to, the true
modalities, and the pipeline needs considerably more samples to resolve the residual ambiguity
(Section~\ref{sec:sample-regularity}). Incorporating a structure term --- e.g.\ preferring clusters whose
members lie on a low-dimensional manifold --- is a natural extension.

\emph{Greediness and seeds.} The assignment is a greedy min-max over a random seed set. There are
no optimality guarantees, and the result depends on the seed set. A bad seed set can produce
unbalanced or impure clusters (size range $[157,386]$ at $K=10$). A deterministic or
seed-robust variant would strengthen the method.

\emph{The $g$ bottleneck.} At inference the branch is chosen by $g(x)$, a function of $x$ alone.
When the same $x$ genuinely supports several modalities, no such function can be correct. This
irreducible error is inherited by the whole pipeline and is visible in the persistent gap between
the DQC test minMSE and its own training MSE.

\emph{Synthetic scope.} All experiments use synthetic data with known ground truth. The behavior
on real multi-modal datasets, where the number of modalities is unknown and the contradiction
geometry is less clean, remains to be tested.

\section{Advantages and Practical Role}
\label{sec:advantages}

The limitations above are the price of the method's distinctive properties, which we now make
explicit.

\emph{Hard, gradient-free, deterministic comparison.} Algorithm~\ref{alg:cluster} uses no learned
similarity, no threshold, no iterative optimization, and no loss: the only quantity is the
geometric ratio $\norm{\Delta y}/\norm{\Delta x}$. The partition is a one-shot, data-only
computation that is frozen before any network is trained.

\emph{Trivially parallel.} Every pairwise contradiction evaluation is an independent norm
computation. Given the current cluster state, the scores of a sample against all candidate
clusters can be computed concurrently, and the entire sweep over samples can be parallelized over
threads. This is an advantage of hard geometric comparisons over learned, iterative procedures,
which are serial by construction.

\emph{Complexity.} The total number of pairwise evaluations is
$\sum_{i=1}^{n}(i-1)=\tfrac{n(n-1)}{2}\approx O(n^2/2)$: the first sample requires no comparison,
the last sample is compared against the $n-1$ samples assigned before it, and the average cost is
quadratic with a factor $1/2$ (in big-$O$ notation, $O(n^2)$). For the sizes considered here the
clustering runs in seconds on a single thread and is far cheaper than training a single network.

\emph{A fast front-end for generative refinement.} The empirical conclusion of this paper is that
difference-quotient-based clustering alone provides \emph{coarse} accuracy --- test minMSE within
$2\times$ of the oracle, but not oracle-level. This is exactly the regime in which the method is
most useful as a \emph{preprocessing} step: it attaches discrete, structurally meaningful
conditional labels to the data at negligible cost, so that a downstream fine-grained generative
model --- e.g.\ flow matching \cite{lipman2022flow} or a diffusion model \cite{ho2020ddpm} ---
does not have to discover the branching structure from scratch but only to refine it. In
multi-modal regression pipelines where the expensive part is the conditional generative model,
spending $O(n^2/2)$ cheap comparisons up front can substantially reduce the training burden of the
refinement stage.

\section{Conclusion and Future Work}
\label{sec:conclusion}

We argued that multi-possible outputs are, at their core, pairwise contradictions between samples,
and we formalized the contradiction as the ratio of output distance to input distance. We tested
whether clustering by this contradiction directly resolves the mean-regression problem, and found
that a difference-quotient clustering, followed by a
logits generator and a conditional network, removes most of the mean collapse: on both $K=5$ and
$K=10$ benchmarks the pipeline reaches test minMSE within $1.4$--$3.2\times$ of an oracle with
true labels, an order of magnitude better than random labels or mean collapse. We reported two empirical regularities --- deeper
networks resolve larger contradictions, and oracle labels generalize from fewer samples than
equivalent cluster labels --- and we positioned the method as a fast, parallelizable,
$O(n^2/2)$ front-end for coarse conditional-label assignment that lowers the training burden of
subsequent generative refinement.

Future work proceeds along four directions. (i)~Adding a low-dimensional-structure term to the
clustering objective to raise purity. (ii)~Integrating the coarse labels with fine-grained
refinement mechanisms, such as trajectory-consistent threading across time, which have been shown
to pin modal identity at oracle-level accuracy. (iii)~Validating the pipeline on real
multi-modal regression datasets where $K$ must itself be estimated from the data. (iv)~Iterative
re-clustering in output space: the first stage produces a conditional predictor
$f_1(x,\mathrm{onehot}(c))$ whose output $y'=f_1(x,\mathrm{onehot}(c))$ already lies closer to one
of the true branches than the unconditional mean does. One can treat this $y'$ as a new input and
the true output $y$ as the new label, run the same difference-quotient clustering on the pair
$(y',y)$ to obtain refined labels $d$, form the augmented input $z=Z(x,y')$, and train a second
logits generator $g_2(z)$ (predicting $d$) together with a second conditional network
$f_2(z,\mathrm{onehot}(d))$. Because the first stage has already displaced samples toward their
modal branches, the contradictions measured in the second pass are dominated by residual fitting
error rather than by genuine multi-modality, which may yield purer clusters and a smaller minMSE.
We leave this two-stage refinement for future work.

\bibliographystyle{plain}

\begin{thebibliography}{9}

\bibitem{bishop2006pattern}
C.~M. Bishop.
\newblock \emph{Pattern Recognition and Machine Learning}.
\newblock Springer, 2006.

\bibitem{bishop1994mdn}
C.~M. Bishop.
\newblock Mixture density networks.
\newblock Technical Report NCRG/94/004, Aston University, 1994.

\bibitem{rupprecht2017uncertain}
C.~Rupprecht, I.~Laina, R.~Dippel, M.~Wimmer, and F.~Tombari.
\newblock Learning in an uncertain world: Representing ambiguity through multiple hypotheses.
\newblock In \emph{ICCV}, 2017.

\bibitem{makansi2019mdn}
O.~Makansi, E.~Ilg, O.~Cicek, and T.~Brox.
\newblock Overcoming limitations of mixture density networks: A sampling and fitting framework for
multimodal future prediction.
\newblock In \emph{CVPR}, 2019.

\bibitem{jacobs1991moe}
R.~A. Jacobs, M.~I. Jordan, S.~J. Nowlan, and G.~E. Hinton.
\newblock Adaptive mixtures of local experts.
\newblock \emph{Neural Computation}, 3(1):79--87, 1991.

\bibitem{ho2020ddpm}
J.~Ho, A.~Jain, and P.~Abbeel.
\newblock Denoising diffusion probabilistic models.
\newblock In \emph{NeurIPS}, 2020.

\bibitem{lipman2022flow}
Y.~Lipman, R.~T.~Q. Chen, H.~Ben-Hamu, M.~Nickel, and M.~Le.
\newblock Flow matching for generative modeling.
\newblock In \emph{ICLR}, 2023.

\end{thebibliography}

\end{document}